\documentclass[10pt,twocolumn,letterpaper]{article}

\usepackage{multirow}
\usepackage[pagenumbers]{cvpr} % To force page numbers, e.g. for an arXiv version

\usepackage[dvipsnames]{xcolor}

\definecolor{cvprblue}{rgb}{0.21,0.49,0.74}
\usepackage[pagebackref,breaklinks,colorlinks,citecolor=cvprblue]{hyperref}

\def\paperID{110} % *** Enter the Paper ID here
\def\confName{CVPR}
\def\confYear{2024}

\title{Improving Depth Gradient Continuity in Transformers: A Comparative Study on Monocular Depth Estimation with CNN}

\author{Jiawei Yao$^1$ \quad Tong Wu$^{1}$ \quad Xiaofeng Zhang$^{2}$\\
$^1$ University of Washington \quad
$^2$ Shanghai Jiao Tong University\\
{\tt\small \{jwyao, tw96\}@uw.edu, framebreak@sjtu.edu.cn}
} 

\begin{document}
\maketitle
\begin{abstract}
Monocular depth estimation is an ongoing challenge in computer vision. Recent progress with Transformer models has demonstrated notable advantages over conventional CNNs in this area. However, there's still a gap in understanding how these models prioritize different regions in 2D images and how these regions affect depth estimation performance. To explore the differences between Transformers and CNNs, we employ a sparse pixel approach to contrastively analyze the distinctions between the two. Our findings suggest that while Transformers excel in handling global context and intricate textures, they lag behind CNNs in preserving depth gradient continuity. To further enhance the performance of Transformer models in monocular depth estimation, we propose the Depth Gradient Refinement (DGR) module that refines depth estimation through high-order differentiation, feature fusion, and recalibration. Additionally, we leverage optimal transport theory, treating depth maps as spatial probability distributions, and employ the optimal transport distance as a loss function to optimize our model. Experimental results demonstrate that models integrated with the plug-and-play Depth Gradient Refinement (DGR) module and the proposed loss function enhance performance without increasing complexity and computational costs on both outdoor KITTI and indoor NYU-Depth-v2 datasets. This research not only offers fresh insights into the distinctions between Transformers and CNNs in depth estimation but also paves the way for novel depth estimation methodologies.

\end{abstract}    
\section{Introduction}

Monocular depth estimation aims to perceive the depth of each pixel in a 2D image, playing a pivotal role in understanding the three-dimensional spatial construction of scenes. Historically, depth maps were acquired using high-end sensors, but their expensive cost and scene limitations hindered widespread adoption. Consequently, extracting depth information from 2D images using monocular cameras has become a research hotspot.

\begin{figure}[htb]
	\centering
	\includegraphics[width=1.0\columnwidth]{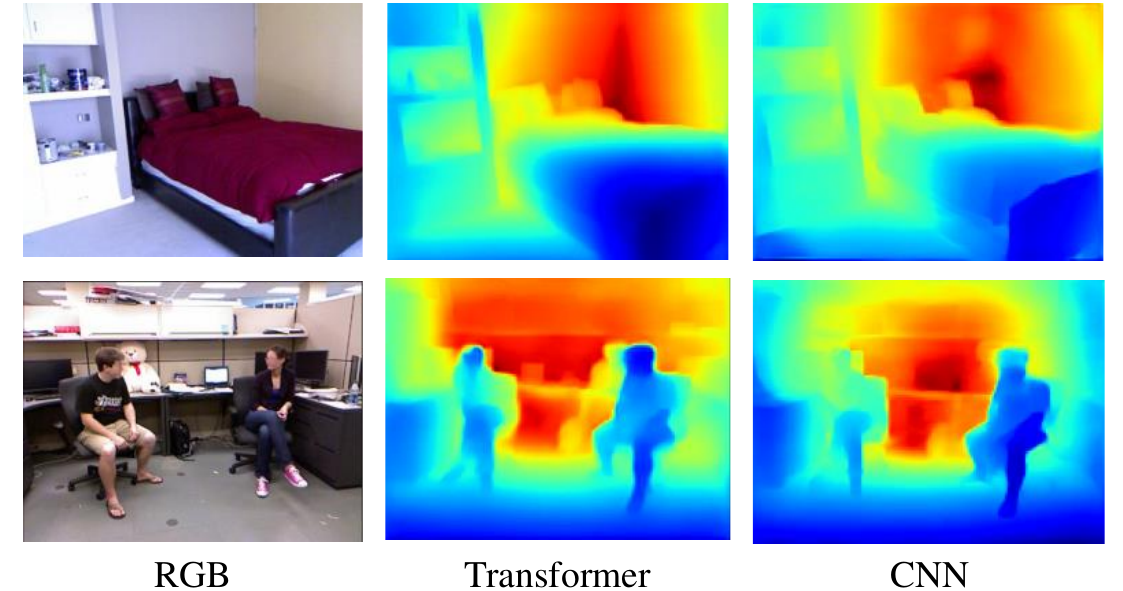}
	\caption{Visualization comparison of depth estimation using Transformers and CNNs. From left to right: RGB, depth prediction using a Transformer encoder, and depth prediction using CNN. Both results are obtained under identical data processing and loss conditions. Depth maps estimated by the Transformer method exhibit clearer scene structures than those by the CNN method, while CNNs provide smoother depth estimations at object boundaries.}
	\label{fig1}

\end{figure}

With the advancement of convolutional networks, backpropagation learning of features has replaced early handcrafted features. Eigen et al. \cite{eigen2014depth} pioneered the use of neural networks for depth feature learning, setting the stage for the rapid development of monocular depth estimation. Hu et al. \cite{hu2019revisiting} built upon CNNs and employed L1 loss, gradient loss, and normal loss to address depth map boundary distortions. Bhat et al. \cite{bhat2021adabins} utilized EfficientNet \cite{tan2019efficientnet} as the encoder and introduced Transformer modules during decoding to enhance global feature correlations, achieving state-of-the-art results. With the success of ViT \cite{dosovitskiy2020image} in image classification, recent works have explored replacing CNNs with Transformers for feature extraction. Yang et al. \cite{yang2021transformer} combined CNNs with Transformers, merging local information from CNNs and global insights from Transformers, offering a novel perspective on global-local information aggregation. Kim et al. \cite{kim2022global} employed Transformers as encoders and introduced a global-local selective fusion module, elevating depth estimation accuracy. Yuan et al. \cite{yuan2022neural} adopted the Swin-Transformer \cite{liu2021swin} as the encoder and leveraged the Transformer's self-attention mechanism with CRF to integrate global information. Rahman et al. \cite{rahman2023dwinformer} introduced DwinFormer, a transformative dual window transformer-based architecture for monocular depth estimation, marking a new benchmark in performance.

Despite Transformers' significant success in monocular depth estimation, outperforming CNN models, the reasons behind their superiority remain an open question. To the best of our knowledge, this issue has not been thoroughly investigated. Delving into this matter will enhance our understanding of Transformers' principles in depth estimation, propelling the field forward. Human vision typically employs cues like size-distance relations, occlusions, and brightness levels to predict object depths in 2D images \cite{torralba2002depth,lebreton2014measuring,saxena2007depth}. This paper seeks to uncover the cues Transformers use for monocular depth estimation and how to amplify these cues to boost performance. We employ visualization techniques to address these questions, examining whether Transformers and CNN models focus on the same regions in 2D images and if these focal points directly influence the experimental outcomes. For visualization, we adopt the method proposed by Hu et al. \cite{hu2019visualization}.

Our experiments demonstrate that Transformers are particularly sensitive to gradient information in images, especially gradients beneficial for scene depth information. However, Transformers lag behind CNNs in handling the continuity of depth gradients. Given these observations, we introduce the Depth Gradient Refinement (DGR) module, a plug-and-play component designed to enhance Transformer performance in depth estimation tasks. Additionally, we propose a loss function optimization based on optimal transport theory. Empirical results confirm that our DGR module and loss function significantly improve the performance of Transformer-based depth estimation models, establishing new state-of-the-art results on popular depth estimation benchmarks. The main contributions of this work are:

\begin{itemize}
    \item We provide a comprehensive comparison between Transformer and CNN models in monocular depth estimation through visualization, offering an interpretable analysis of their focal regions and operational principles.
    \item To address the challenges of depth gradient continuity in Transformers, we introduce the novel Depth Gradient Refinement (DGR) module. This paper also presents a unique perspective by treating depth maps as spatial probability distributions and employs optimal transport distance as a loss function for model optimization.
    \item Our proposed method, designed as a plug-and-play component, seamlessly integrates with existing Transformer-based monocular depth estimation models. When combined with leading Transformer-based models, our approach achieves breakthrough performance, surpassing existing benchmark.
\end{itemize}
\section{Related Work}

\subsection{CNNs and Transformers for Monocular Depth Estimation}
With the evolution of neural networks, the translational invariance \cite{wang2020cnn} and robust feature representation capabilities of CNNs have been progressively employed as backbone networks in depth estimation, object detection, and semantic segmentation. However, due to the constraints of their receptive field size, CNNs have struggled with long-range dependencies until the introduction of Transformers in the image domain. By adopting a sequence-based approach, Transformers possess a global receptive field. With continuous refinements, they have emerged as a dominant network architecture, often outperforming CNNs. Zhang et al. \cite{zhang2023lite} introduced a lightweight hybrid of CNNs and Transformers for self-supervised monocular depth estimation, incorporating a Consecutive Dilated Convolutions (CDC) module and a Local-Global Features Interaction (LGFI) module. Shim and Kim \cite{shim2023swindepth} introduced SwinDepth, utilizing the convolution-free Swin Transformer, while Rahman et al. \cite{rahman2023dwinformer} presented DwinFormer, a dual window transformer-based architecture. Despite the superior performance of Transformer-based models, they often lack the fine-grained filtering capabilities of CNNs for detailed information \cite{hassani2021escaping}.
\begin{figure*}[htb]
	\centering
	\includegraphics[width=0.9\textwidth]{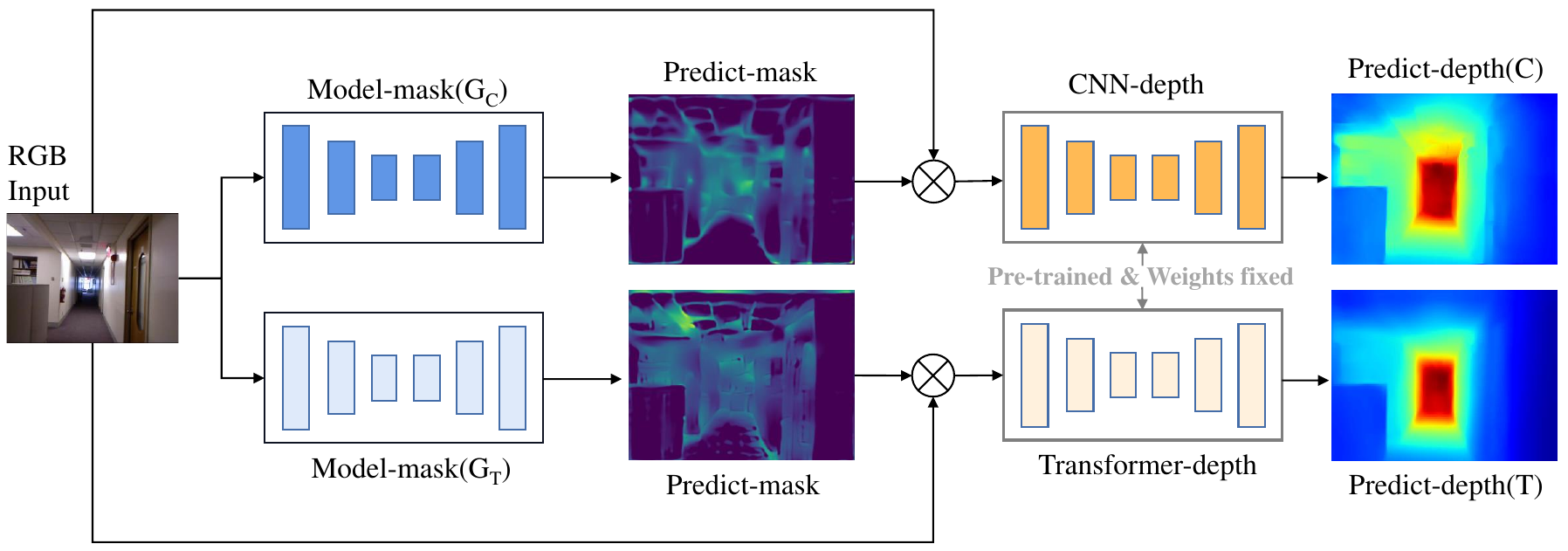}
	\caption{Network architecture for visualizing Depth Estimation differences between CNN and Transformer. This figure illustrates a dual-pathway architecture for depth estimation from RGB input. It includes two parallel branches with model-mask networks ($G_C$ and $G_T$) generating predictive masks from the input image. The masks are then element-wise multiplied with the RGB input to isolate features of interest. The resulting representations are fed into pre-trained and weight-fixed depth prediction networks (“CNN-depth” and “Transformer-depth”), which independently predict depth maps (Predict-depth(C) and Predict-depth(T)). The model-masks networks are trained to mask out regions that are not of interest to the depth prediction models, enhancing the alignment of the predicted depth maps with the GTs.}
	\label{fig2}

\end{figure*}

\subsection{Visualization in Deep Learning Models}
Interpretability in deep learning has garnered significant attention, especially in the sub-domains of image classification and generation. In traditional image classification models, RGB three-channel images are typically inputted, processed through convolutional and downsampling layers, and then passed through fully connected layers to output class probabilities. Due to the mathematical properties of convolution and downsampling, there's a clear spatial correspondence between feature maps and input images. This has led to the common practice of using feature map activations or gradients for input image visualization.
In recent years, numerous studies have aimed to unveil the inner workings of CNNs in image classification. For instance, Selvaraju et al. \cite{selvaraju2017grad} employed gradient-based methods, which explore model output variations by modifying parts of the input image. Class Activation Mapping (CAM) has proven effective in classification tasks, computing the last layer activations as linear combinations across channels. However, CAM has limitations, especially for complex network structures. Grad-CAM was introduced to integrate gradient information, addressing CAM's shortcomings. To achieve more detailed visualizations, researchers further introduced Guided Backpropagation, enhancing GradCAM's visualization with higher resolution and clearer semantics.
However, when considering Transformers, there's a noticeable scarcity of interpretability studies in the visual domain. Abnar et al.~\cite{abnar2020quantifying} hypothesized linear relationships in attention points, considering attention graph paths and constructing input saliency maps using self-attention during forward propagation. Yet, it fails to reflect the decision-making impact of different categories due to potential attenuation of average attention scores. Chefer et al.~\cite{chefer2021transformer} allocated local correlations based on deep Taylor decomposition, subsequently propagating these correlations across layers, involving both attention and residual connections.
It's crucial to note that most aforementioned methods are designed for image classification, making them unsuitable for depth estimation visualization. Depth estimation fundamentally differs from classification: its output is a 2D depth map, not class probabilities. This suggests potential vast differences in feature focal points between the two. Previous work by Hu et al. \cite{hu2019visualization} provided insights into how CNNs select depth cues from 2D images. However, the mechanisms behind Transformers remain enigmatic. Moreover, while Transformers often outperform CNNs under identical data processing, the reasons behind this superiority warrant further exploration.
\section{Methodology}

We adopt the visualization approach proposed by Hu et al~\cite{hu2019visualization}. for the interpretability experiments in monocular depth estimation. The underlying assumption is that the Transformer network can extract depth information from a selected set of pixels. If a subset of pixels in an image can approximate the entire image with results within an acceptable range, we can identify the regions of interest for the network. By analyzing the commonalities among these regions, we can deduce the cues the Transformer network uses for depth prediction. Building on this, we introduce the Depth Gradient Refinement (DGR) module and the Optimal Transport Depth Loss (OTDL).

\begin{figure}[htb]
	\centering
	\includegraphics[width=1\linewidth]{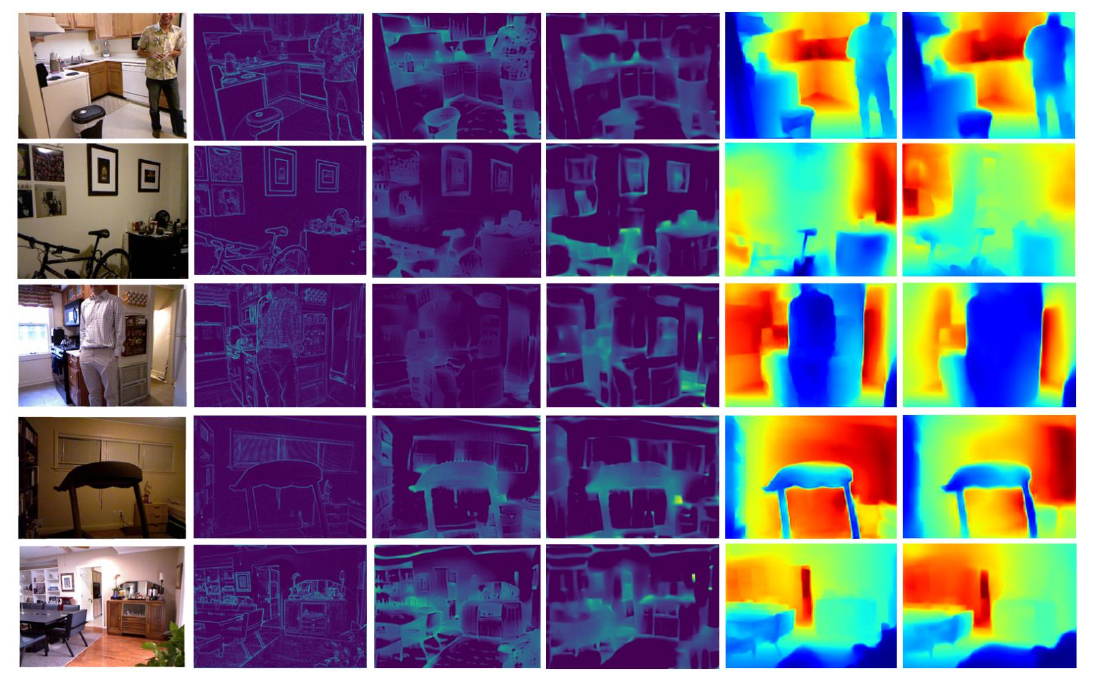}
	\caption{Visualization of differences. From left to right: RGB, image gradient, masks predicted by Transformer and CNN, and depth predictions post-sparsification.}

	\label{fig3}
\end{figure}

\subsection{Visualization of Monocular Depth Estimation}
  The overall architecture of our experimental network is illustrated in Figure~\ref{fig2}. For our study, we select two networks, namely $N_{t}$ (Transformer) and $N_{c}$ (CNN), to predict the depth maps of a single image, denoted as $Y_{t}$ and $Y_{c}$, respectively. The input image is represented as $I$. During the training phase of both networks, we employ consistent data processing techniques and utilize the same loss function.
\begin{equation}
	Y_{t} = N_{t}(I)
\end{equation}
\begin{equation}
	Y_{c} = N_{c}(I)
\end{equation}

The aforementioned approach involves feeding the entire image into the network. To investigate which regions of the image the network prioritizes, we employ a sparse pixel mask, denoted as $M$, to occlude certain areas of the input image. If the network can still accurately predict the depth of the partially occluded image, then the unmasked portions of the image are deemed as the regions of interest for the network. We define the depth predictions of the Transformer and CNN for the sparse image as $\hat{y}_{t}$ and $\hat{y}_{c}$, respectively.
\begin{equation}
	\hat{y}_{t} = N_{t}(I*M_{t})
\end{equation}
\begin{equation}
	\hat{y}_{c} = N_{c}(I*M_{c})
\end{equation}
where $M_{t}$ and $M_{c}$ represent the sparse pixels selected by the Transformer and CNN networks, respectively. If an appropriate mask $M$ is chosen such that $\hat{y}$ approximates $Y$ (i.e., $Y_{t} \approx \hat{y}_{t}$ and $Y_c \approx \hat{y}_{c}$). We process the mask in two distinct ways: first, by binarizing it, ensuring all elements in the mask are either 0 or 1; second, by allowing the mask values to be continuous values ranging between 0 and 1. In our experiments, we aim to identify the cues Transformers use in images and compare them with the findings of \cite{hu2019visualization} to discern the differences with CNNs. Ultimately, we formulate the problem as the following optimization:
\begin{equation}
	\mathop{\min}_{M} L_{dif}(Y,\hat{Y}) + \lambda \frac{1}{n}||M||_1
\end{equation}
where $L_{dif}$ computes the loss between the depth predictions obtained from the full image and those from the sparse image. $n$ denotes the total number of sparse pixels, $||M||_1$ represents the $L1$ regularization of $M$, and $\lambda$ is a hyperparameter used to control the sparsity level.

We employ two additional networks, $G_t$ and $G_c$, to predict two sets of sparse pixels, $mask_t$ and $mask_c$, respectively. More specifically, we consider the following optimization:
\begin{equation}
	\mathop{\min}_{G} L_{dif}(Y,N(I*G(I))) + \lambda \frac{1}{n}||G(I)||_1
\end{equation}

For network $G$, we constrain its output by appending a sigmoid activation function, ensuring its range lies between 0 and 1. Thus, $I*G(I)$ represents a weighted selection of the input image $I$. Through learning, the network can assign smaller weights to less significant regions, retaining areas it deems important.

To compare the regions of interest for CNN and Transformer, we input the regions selected by the CNN (i.e., $I*mask_c$) into the Transformer and observe the Root Mean Square Error (RMSE). Similarly, we input $I*mask_{t}$ into the CNN to observe its RMSE. If the error rates of the two networks do not exhibit a significant increase, we infer that both networks focus on similar regions. A pronounced difference in error rates would indicate distinct areas of interest between the two networks.
\begin{equation}
	Y_t = N_t(I*mask_c)
\end{equation}
\begin{equation}
	Y_c = N_c(I*mask_t)
\end{equation}
\begin{figure*}[t]
	\centering
	\includegraphics[width=0.9\textwidth]{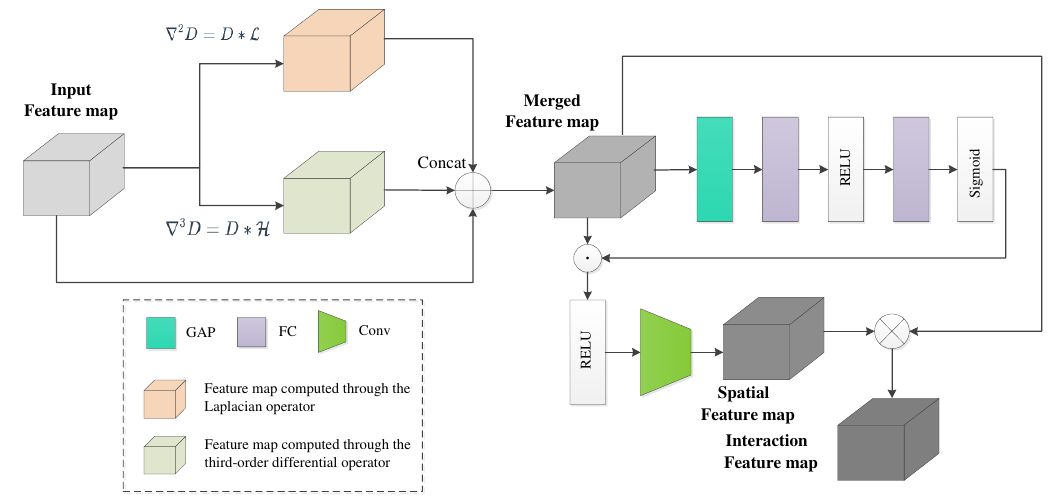}
	\caption{Schematic of the Depth Gradient Refinement (DGR) module. Features processed by the transformer encoder serve as inputs to the DGR module, subsequently undergoing higher-order derivative computation, feature concatenation, and feature recalibration.}

	\label{fig4}
\end{figure*}
Specifically, the masks learned by the two networks, $mask_c$ and $mask_t$, are binarized, meaning each element in the mask is either 1 or 0. Consequently, the product (mask*image) yields the image post-sparsification. We then subject both networks to experimental exploration within each other's regions of interest, aiming to discern the similarities and differences between the Transformer and CNN.

Through extensive experiments on the NYU-depth-V2~\cite{silberman2012indoor} indoor dataset, we elucidate how the Transformer extracts features and its distinctions from the CNN. As depicted in Figure~\ref{fig3}, our experiments reveal that the Transformer model exhibits a pronounced interest in the boundaries (depth gradients) of objects within images, especially the peripheral regions of objects. Such boundaries typically facilitate depth perception in scenes. However, when it comes to handling the continuity of depth gradients, the Transformer lags behind the CNN. Specifically, the depth maps generated by the former might exhibit unnatural depth transitions in certain smooth areas. Detailed experimental results (Table 1) and analyses are presented in Transformer Estimation Depth from Sparse Pixels part.

\subsection{Depth Gradient Refinement Module}

In the realm of monocular depth estimation, transformer models excel at capturing global context but tend to underperform in regions with significant gradient changes, such as object edges. This can lead to depth estimations that are less sharp and less distinct at the boundaries between objects and their surroundings. To tackle this issue, we introduce the Depth Gradient Refinement (DGR) module to enhance the continuity of depth estimation in transformers. Initially, we contemplate employing higher-order derivatives to capture intricate variations in the depth map, especially in regions where object boundaries exhibit rapid intensity transitions. The second and third-order derivatives of the input depth map $D$ are computed as:
\begin{equation}
	\nabla^2 D = D * \mathcal{L}
\end{equation}
\begin{equation}
	\nabla^3 D = D * \mathcal{H}
\end{equation}
where $\mathcal{L}$ denotes the Laplacian operator, and $\mathcal{H}$ represents the corresponding third-order differential operator. The symbol ``$*$" signifies the convolution operation.

Subsequently, leveraging feature concatenation, we amalgamate these higher-order derivative features with the original depth features, yielding a feature representation enriched with multifaceted depth information:
\begin{equation}
	F_{\rm merged} = \rm{Concat}(F_D, \nabla^2 D, \nabla^3 D)
\end{equation}

Lastly, a feature recalibration submodule is employed to further optimize this merged feature, tailoring it for depth estimation tasks. Specifically, a channel attention mechanism is invoked to weight the merged features based on the significance of each channel. Spatial adaptivity is then introduced to bolster feature continuity spatially. To account for higher-order feature interactions, an outer product followed by dimensionality reduction is incorporated. The resultant feature map ensures effective utilization and amplification of information gleaned from both the higher-order derivative features and the original ones, offering a richer context for monocular depth estimation. The computational formula for the feature recalibration submodule is:
\begin{equation}
	w_c = \sigma\left( \rm{FC}_2\left( \rm{ReLU}\left( \rm{FC}_1\left( \rm{GAP}(\textit{F}_{\rm{merged}}) \right) \right) \right) \right)
\end{equation}
\begin{equation}
	F_{{\rm spatial}} = { \rm Conv}\left( {\rm ReLU}\left( {F}_{{\rm merged}} \odot {w}_{c} \right) \right)
\end{equation}
\begin{equation}
	F_{\rm{interaction}} = \rm{Conv}_{1x1}\left( \textit{F}_{\rm{spatial}} \otimes\textit{F}_{\rm{merged}} \right)
\end{equation}

The DGR module is strategically positioned after each encoder block of the Transformer. This design choice stems from the rationale that, on one hand, it facilitates the progressive refinement of depth features, and on the other, each encoder block in the Transformer, encompassing self-attention mechanisms and feed-forward neural networks, might induce feature discontinuities. By introducing the DGR post each block, we ensure continuity is maintained throughout the network.

\subsection{Optimal Transport Depth Loss}

Transformers, devoid of local convolution operations, can occasionally produce depth maps with unnatural depth jumps in regions expected to be smooth. This observation propels the need for a more nuanced loss function that can address this continuity challenge. To bridge this gap, we introduce the Optimal Transport Depth Loss (OTDL). Drawing inspiration from the optimal transport theory, this loss offers a meticulous comparison between predicted and true depth maps, emphasizing the preservation of depth distribution variance and continuity.

To begin with, depth maps must be represented as normalized distributions. Given a predicted depth map $P$ and its corresponding ground truth depth map $Q$, normalization is carried out as:
\begin{equation}
	P^{\prime}=\frac{P}{\sum_{i,j}P(i,j)}
\end{equation}
\begin{equation}
	Q^{\prime}=\frac Q{\sum_{i,j}Q(i,j)}
\end{equation}
where $P^{\prime}$ and $Q^{\prime}$ are interpreted as probability distributions.

Central to optimal transport is the cost matrix, which details the ``expense" of transporting ``mass" between positions. In the depth map context, the depth values provide this positional information. Thus, our cost matrix $M$ has entries:
\begin{equation}
	M_{ij}=|i-j|^2
\end{equation}
with $i$ and $j$ being depth values. The matrix entry $M_{ij}$ denotes the cost of transitioning from depth $i$ to depth $j$.

The core of our proposed loss rests on solving the optimal transport problem:
\begin{equation}
	OT(P^{\prime},Q^{\prime})=\min_{T\in\Pi(P^{\prime},Q^{\prime})}\sum_{i,j}T_{ij}M_{ij}
\end{equation}
where $T$ signifies a joint distribution such that the marginals of $T$ align with $P^{\prime}$ and $Q^{\prime}$. The ensemble $\Pi(P^{\prime},Q^{\prime})$ contains all feasible $T$ distributions that fulfill this criteria.

From the above discussions, the Optimal Transport Depth Loss (OTDL) is articulated as:
\begin{equation}
	L_{\mathrm{OTDL}}(P,Q)=OT(P^{\prime},Q^{\prime})
\end{equation}

For monocular depth estimation, the conventional Mean Squared Error (MSE) loss, denoted as $L_{\mathrm{MSE}}$ , is typically employed:
\begin{equation}
	L_{\mathrm{MSE}}(P,Q)=\frac1N\sum_{i=1}^N(P_i-Q_i)^2
\end{equation}

To harness both the global depth estimation capability of MSE and the depth distribution preservation of OTDL, we combine them to formulate the final Loss:
\begin{equation}
	L(P,Q)= L_{\mathrm{MSE}}(P,Q)+\lambda_{OTDL}\cdot L_{\mathrm{OTDL}}(P,Q)
\end{equation}
where $\lambda_{OTDL}$ is the hyperparameter that regulates the influence of the respective loss components.
\section{Experiments}

\begin{figure}[t]
	\centering
	\includegraphics[width=0.8\columnwidth]{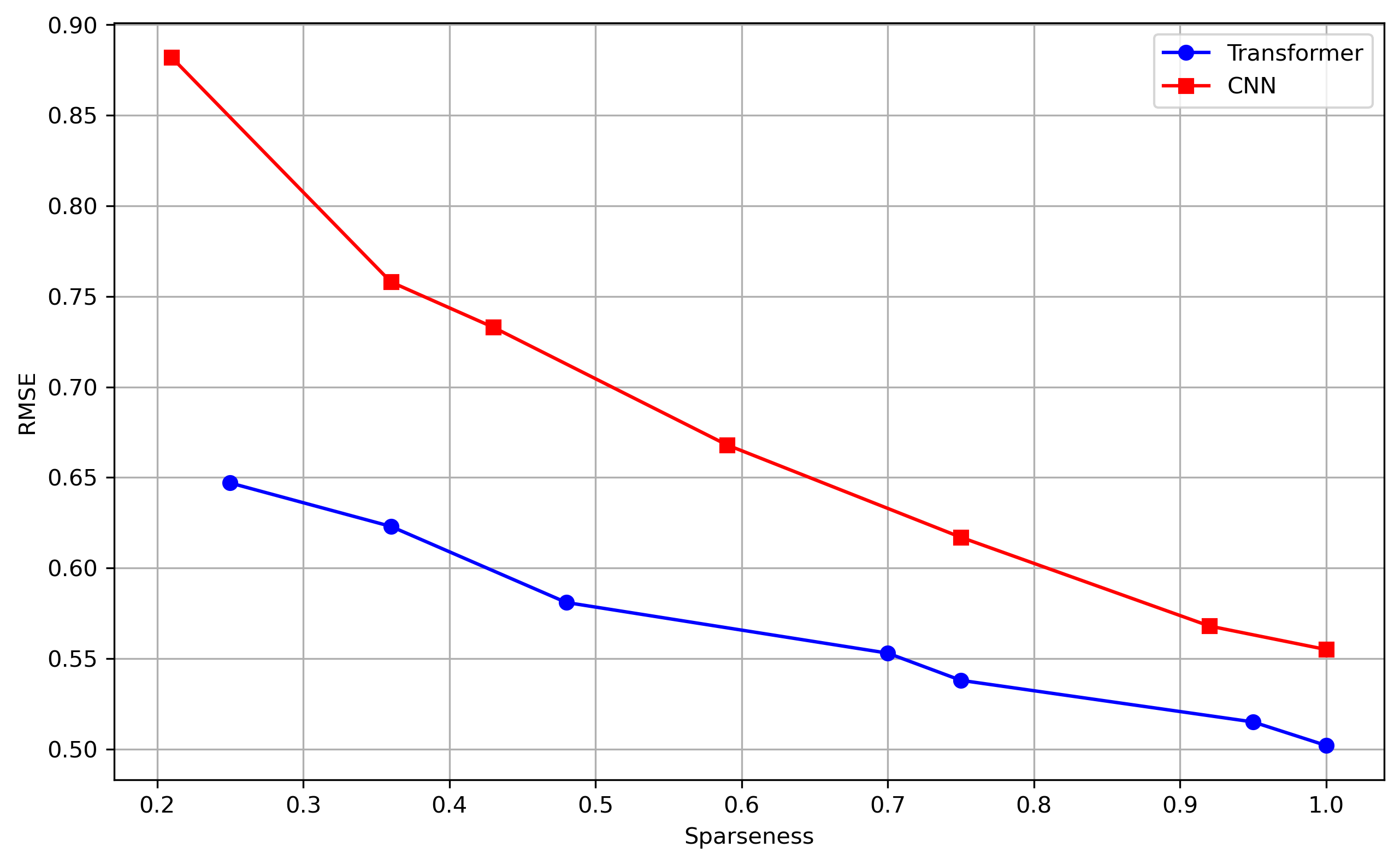}
	\caption{Performance of the two networks at different sparsity levels. A larger RMSE indicates poorer performance.}
	\label{fig5}
\end{figure}

\begin{figure*}[htb]
	\centering
	\includegraphics[width=1\textwidth]{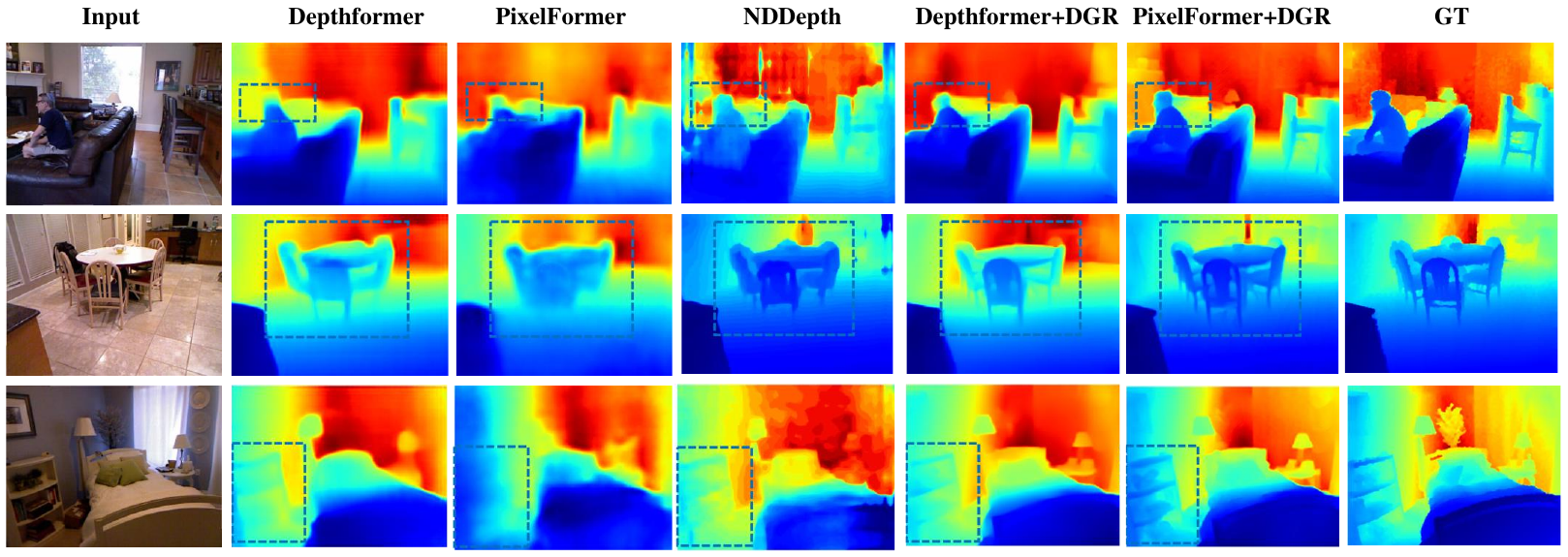}
 
	\caption{Qualitative comparison of different models on indoor dataset NYU-Depth-V2.}
 
	\label{fig:NYU}
\end{figure*}

\begin{table}[]
	
	\centering
	\resizebox{0.9\columnwidth}{!}{%
			\begin{tabular}{c|cccc}
			\hline
			$\lambda$& RMSE & Sparseness & RMSE & Sparseness \\ \hline
			original& 0.502 & 1.00 & 0.555 & 1.00 \\
			$\lambda=1$& 0.515 & 0.95 & 0.568 & 0.92 \\
			$\lambda=2$& 0.538 & 0.75 & 0.617 & 0.75 \\
			$\lambda=3$& 0.553 & 0.70 & 0.668 & 0.59 \\
			$\lambda=4$& 0.581 & 0.48 & 0.733 & 0.43 \\
			$\lambda=5$& 0.623 & 0.36 & 0.758 & 0.36 \\
			$\lambda=6$& 0.647 & 0.25 & 0.882 & 0.21 \\ \hline
		\end{tabular}
	}
	\caption{Pilot study results on NYU-Depth-V2. Depth reconstruction performance of two networks at different sparsity levels. From left to right, the columns represent the hyperparameter controlling sparsity, depth estimation accuracy of Transformer and CNN, sparsity level, and RMSE based on ground truth predictions.}

    \label{tab: tab1}
\end{table}

\textbf{Implementation Details} Our proposed method is implemented using the PyTorch framework on an RTX3090 GPU. We train our model for 200 epochs with a patch size of 256×256. The Adam optimizer is employed with parameters $\beta_1=0.9$, $\beta_2=0.999$, and $\epsilon=1e^{-6}$. The weight decay factors for the encoder and decoder are set to 0.01 and 0, respectively. We adopt a polynomial decay for learning rate scheduling, starting with an initial rate of $10^{-4}$ and a power of $p=0.9$, decaying until the rate reaches $10^{-5}$. For the NYU-Depth-V2 dataset, the input/output resolution during training is set to 416×544.

\paragraph{Datasets} We conduct training and visualization evaluations of Transformer and CNN on the NYU-Depth-V2 dataset~\cite{silberman2012indoor} and assess the performance of our proposed method on this dataset. Then, we evaluated the proposed method on the NYU-Depth-V2 and KITTI datasets.

\paragraph{Models} To delve into the differences, we select the Resnet50 \cite{he2016deep} and SegFormer \cite{xie2021segformer} network models as our target models. For a fair comparison, both models are implemented under identical data processing, training loss functions, and iteration cycles. For the loss function of the target networks, we employ the loss function proposed in \cite{hu2019revisiting} for training.
\begin{equation}
	L_{dif} = L_{depth} + L_{grad} + L_{normal}
\end{equation}
\begin{equation}
	Loss = \frac{1}{n}\sum_{i}d_i^2-\frac{1}{2n^2}\sum_id_i^2
\end{equation}
where $L_{depth} = \frac{1}{n}\sum_{i=1}^nF(e_i)$, $L_{normal} = \frac{1}{n}\sum_{i=1}^n(1-cos\partial_i)$,  $L_{grad}=\frac{1}{n}\sum_{i=1}^n(F(\nabla_x(e_i))+F(\nabla_y(e_i)))$, $F(e_i)=\ln(e_i + 0.5))$.

To assess the efficacy of DGR, we embed it into the state-of-the-art Transformer-based monocular depth estimation models for evaluation.
\begin{table}[]
	\centering
	\resizebox{1\columnwidth}{!}{%
			\begin{tabular}{c|c|cccc}
			\hline
			Method & Venue & Abs Rel$\downarrow$ & RMS$\downarrow$ & ${\rm Log}_{10}\downarrow$ & $\delta_1\uparrow$ \\ \hline
			DORN~\cite{fu2018deep}	& CVPR’18 & 0.115 & 0.509 & 0.051 & 0.828 \\
			Yin et al.~\cite{yin2019enforcing} & ICCV’19	& 0.108 & 0.416 & 0.048 & 0.872 \\
			Adabins~\cite{bhat2021adabins} & CVPR’21	& 0.103 & 0.364 & 0.044 & 0.903 \\
			DPT~\cite{ranftl2021vision}	& ICCV’21 & 0.110 & 0.367 & 0.045 & 0.904 \\
			TransDepth~\cite{yang2021transformer} & ICCV’21	& 0.106 & 0.365 & 0.045 & 0.900 \\
			SwinDepth~\cite{cheng2021swin} & IEEE SENS J’21	& 0.100 & 0.354 & 0.042 & 0.909 \\
			DepthFormer~\cite{agarwal2022depthformer}	& ICIP’22 & 0.100 & 0.345 & - & 0.911 \\ 
            NeWCRFs~\cite{yuan2022neural} & CVPR’22 & 0.095 & 0.334 & 0.041 & 0.922 \\
			PixelFormer~\cite{agarwal2023attention}	& WACV’23 & 0.090 & 0.322 & 0.039 & 0.929 \\

            NDDepth~\cite{shao2023nddepth} & ICCV’23 & 0.087 & 0.311 & 0.038 & 0.936\\

            IEBins~\cite{shao2023iebins} & Arxiv’23 & 0.087 & 0.314 & 0.038 & 0.936\\

            \hline
			
			Adabins + DGR & Ours & 0.097 & 0.347 & 0.041 & 0.918 \\
			DPT + DGR	& Ours	& 0.104 & 0.348 & 0.042 & 0.914 \\
			TransDepth + DGR & Ours		& 0.101 & 0.348 & 0.043 & 0.911 \\
			SwinDepth + DGR	& Ours	& 0.094 & 0.336 & 0.040 & 0.920 \\
			DepthFormer + DGR	& Ours	& 0.096 & 0.329 & - & 0.922 \\
			PixelFormer + DGR	& Ours	& \textcolor{blue}{\textbf{0.086}} & \textcolor{blue}{\textbf{0.310}} & \textcolor{blue}{\textbf{0.036}} & \textcolor{blue}{\textbf{0.937}} \\ \hline
		\end{tabular}
	}
	\caption{Experimental results on NYU-Depth-V2. Bold text indicates the best performance.}
 \label{tab:nyu}

\end{table}
\subsection{Transformer Estimation Depth from Sparse Pixels}
As shown in Table~\ref{tab: tab1} and Figure~\ref{fig5}, we progressively increase the sparsity level of the input image. Our observations are as follows: (1) The depth reconstruction accuracy of both networks decreases with increasing mask sparsity. At the same sparsity level, the performance of the Transformer is notably superior to that of the CNN. Remarkably, the Transformer, when recognizing only 25\% of the image regions, exhibits comparable performance to the CNN recognizing 60\% of the regions. (2) The Transformer demonstrates better robustness compared to the CNN. As evident from the table, with increasing sparsity, even when retaining only 36\% of the input image information (the sparsity level used in subsequent experiments), the RMSE drops by only 0.123 for the Transformer, while it drops by 0.203 for the CNN.

From the above data analysis, it is evident that the Transformer exhibits better robustness than the CNN. However, the exact nature of this disparity remains elusive. To delve deeper, we conduct a visual analysis at a sparsity level of 36\% for both models.
\begin{figure}[t]
	\centering
	\includegraphics[width=\linewidth]{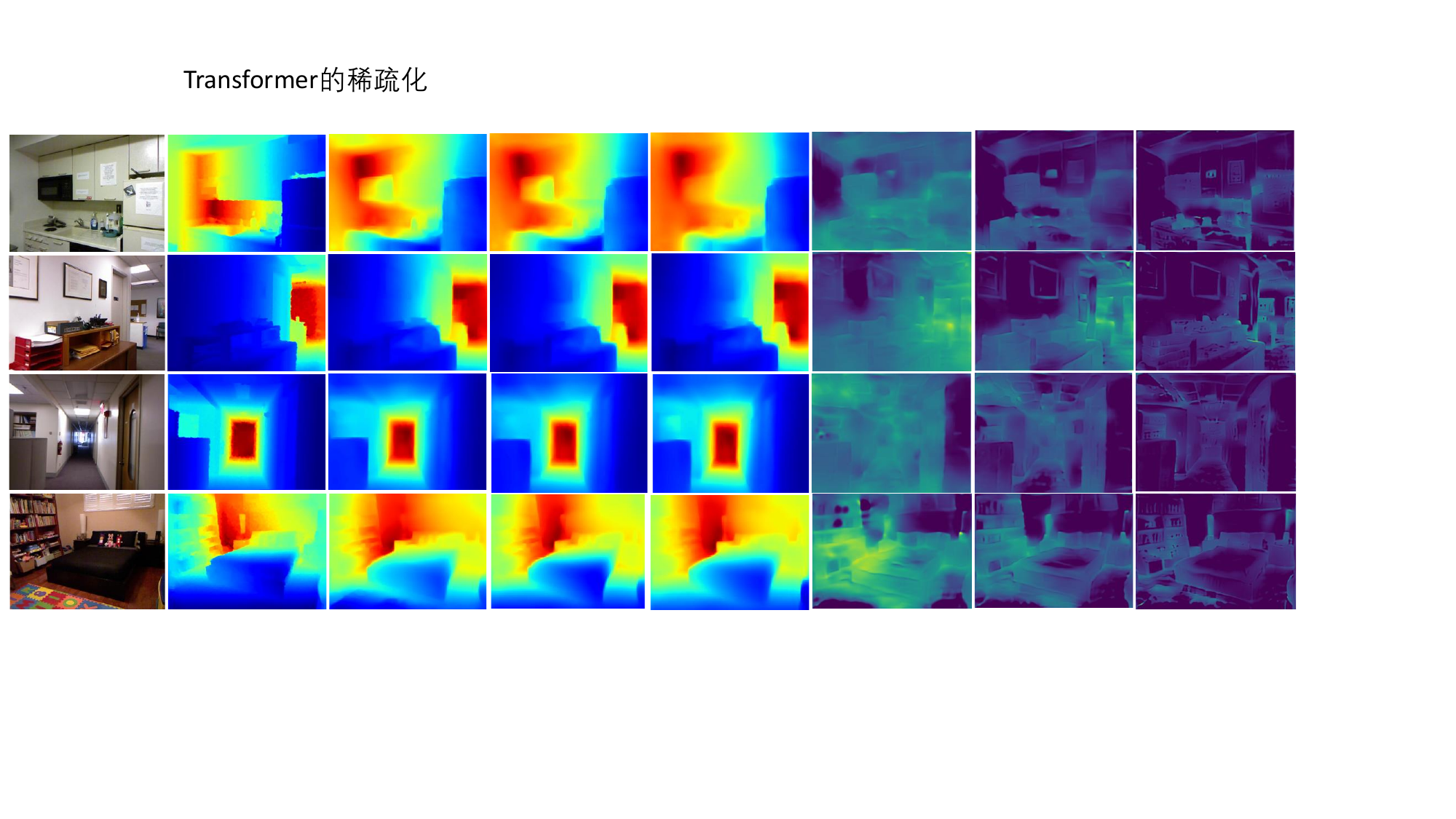}
 
	\caption{From left to right: RGB, Ground Truth, depth maps predicted by the network, and areas of interest in the image when $\lambda$ = 1, 3, 5.}
	\label{fig6}

\end{figure}
To further investigate the differences between the two networks, we analyze their differences through subjective visual results. Figure~\ref{fig6} displays the input image, masks trained by Transformer and CNN, and depth maps obtained from sparse pixels by Transformer and CNN. From the images, we observe that: (1) Both Transformer and CNN show interest in similar regions when selecting depth cues. However, the Transformer is more sensitive to image boundaries and object contours than the CNN, resulting in clearer and more accurate depth estimation at the boundaries. Moreover, the Transformer has a stronger ability to distinguish between the foreground and background of an image, whereas the CNN might not differentiate them well in certain scenarios. (2) Due to its global attention mechanism, the Transformer captures the contextual relationships of the entire image better, especially in distant areas. This makes its depth estimation between objects and the background more accurate. In contrast, the CNN, with its convolution operation, excels in capturing local textures and shape information, producing depth maps with smooth depth gradients. However, for complex textures or color patterns, the CNN might misinterpret. (3) While the Transformer captures global information, its generated depth map might show unnatural depth jumps in some smooth areas, affecting continuity. The CNN excels in producing depth maps with clear depth gradients, especially at object edges and in texture-rich areas, offering a more vivid and three-dimensional visual effect.

\subsection{Comparison to previous state-of-the-art competitors}

In this section, we analyze the experimental results obtained by incorporating our proposed DGR (Dynamic Gradient Refinement) module with various Transformer-based state-of-the-art methods for monocular depth estimation. In addition, we compare these results with those from several prominent CNN-based SOTA models. Our experiments span across two benchmark datasets: NYU-Depth-V2~\cite{silberman2012indoor} and KITTI~\cite{geiger2013vision}.

\begin{table}[]
	
	\centering
	\resizebox{1\columnwidth}{!}{%
			\begin{tabular}{c|c|cccc}
			\hline
			Method & Venue & Abs Rel$\downarrow$ & RMS$\downarrow$ & ${\rm Log}_{10}\downarrow$ & $\delta_1\uparrow$ \\ \hline
			DORN~\cite{fu2018deep}	& CVPR’18 & 0.072 & 2.727 & 0.120 & 0.932 \\
			Yin et al.~\cite{yin2019enforcing} & ICCV’19	& 0.072 & 3.258 & 0.117 & 0.938 \\
			Adabins~\cite{bhat2021adabins} & CVPR’21	& 0.058 & 2.360 & 0.088 & 0.964 \\
			DPT~\cite{ranftl2021vision}	& ICCV’21 & 0.062  & 2.573 & 0.092 & 0.959 \\
			TransDepth~\cite{yang2021transformer} & ICCV’21	& 0.064 & 2.755 & 0.098 & 0.956 \\
			SwinDepth~\cite{cheng2021swin} & IEEE SENS J’21	& 0.106 & 4.510 & 0.182 & 0.890 \\
			DepthFormer~\cite{agarwal2022depthformer}	& ICIP’22 & 0.052 & 2.143 &0.079 & 0.975 \\ 
            NeWCRFs~\cite{yuan2022neural} & CVPR’22 & 0.052 & 2.129 & 0.079 & 0.974 \\
			PixelFormer~\cite{agarwal2023attention}	& WACV’23 & 0.051 &2.081 &0.077 & 0.976 \\

            NDDepth~\cite{shao2023nddepth} & ICCV’23 & 0.050 & \textcolor{blue}{\textbf{2.025}} & 0.075 & 0.978\\

            IEBins~\cite{shao2023iebins} & Arxiv’23 & 0.051 & 2.370 & 0.076 & 0.974\\

            \hline
			
			Adabins + DGR & Ours & 0.055 & 2.357 & 0.083 & 0.967 \\
			DPT + DGR	& Ours	& 0.060 & 2.568 & 0.088 & 0.963 \\
			TransDepth + DGR & Ours		&0.061 & 2.748 & 0.094& 0.962 \\
			SwinDepth + DGR	& Ours	& 0.098 & 4.485 & 0.151 & 0.894 \\
			DepthFormer + DGR	& Ours	& 0.050 & 2.124 & \textcolor{blue}{\textbf{0.074}}& 0.979 \\
			PixelFormer + DGR	& Ours	& \textcolor{blue}{\textbf{0.049}} & 2.041 & 0.075 & \textcolor{blue}{\textbf{0.979}} \\ \hline
		\end{tabular}
	}

	\caption{Experimental results on KITTI. Bold text indicates the best performance.}
 
 \label{tab:kitti}
\end{table}

\paragraph{NYU-Depth-V2 Results} Table~\ref{tab:nyu} presents the comparative results on NYU-Depth-V2 dataset. Our proposed Adabins + DGR shows a significant improvement over the baseline Adabins model, reducing the Absolute Relative Difference (Abs Rel) from 0.103 to 0.097 and RMSE from 0.364 to 0.347. This improvement demonstrates the efficacy of the DGR module in refining depth predictions. Notably, PixelFormer integrated with DGR (PixelFormer+DGR) outperforms all other methods, achieving the best performance across most metrics, specifically lowering Abs Rel to 0.086 and RMSE to 0.310. This result underscores the compatibility of our DGR module with different transformer-based architectures, enhancing their depth estimation capabilities. Figure~\ref{fig:NYU} shows the visual comparison on NYU-Depth-V2 dataset. Examining the edge definition, the integration with the DGR module appears to enhance the edge smoothness, particularly around object boundaries. In addition, the detail preservation in areas of intricate geometry, like the patterned chair backs and room corners, is visibly better in the DGR-enhanced models.

\begin{figure}[htb]
	\centering
	\includegraphics[width=1\linewidth]{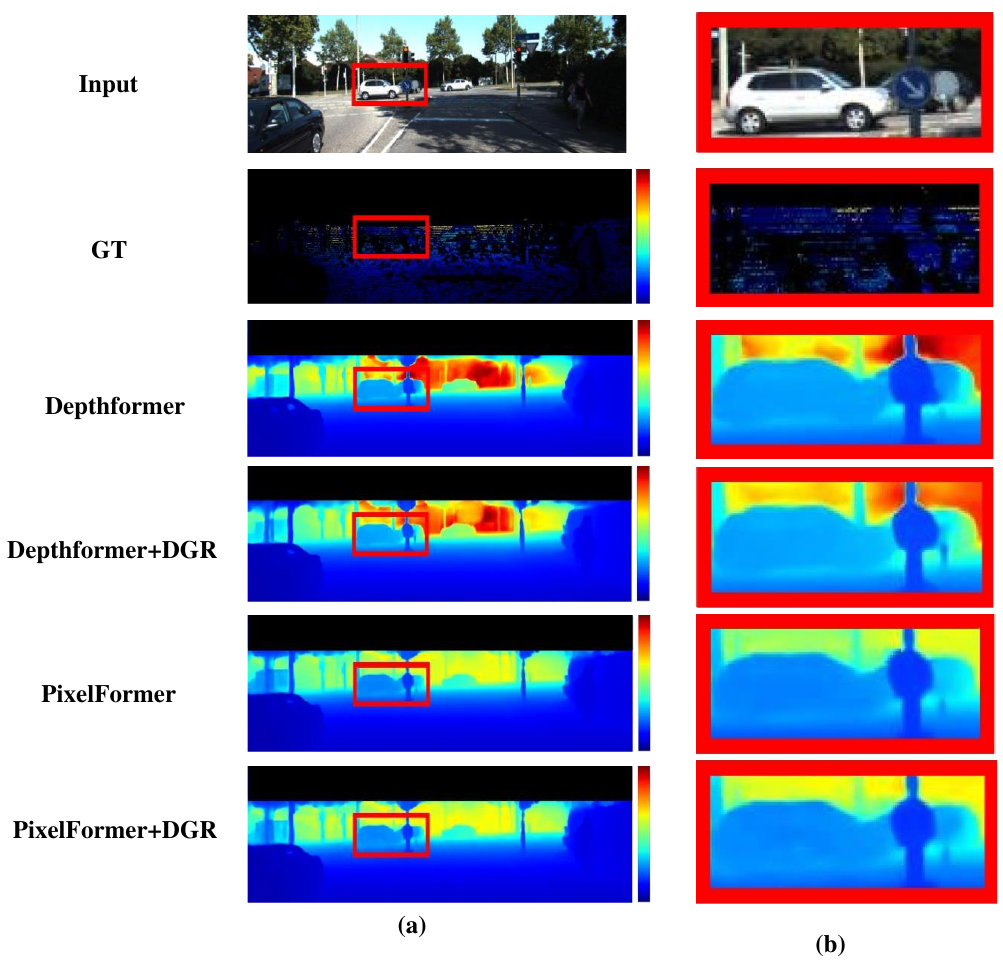}

	\caption{Comparative Visualization of Depth Estimation on KITTI Dataset with DGR Enhancement. (b) is a close-up view of the red frame in (a).}
 
	\label{fig:KITTI}
\end{figure}
\paragraph{KITTI Results} As shown in Table~\ref{tab:kitti}, on KITTI dataset, the Depthformer + DGR and PixelFormer + DGR again demonstrate superior performance, with Depthformer + DGR achieving the lowest Abs Rel of 0.050 and PixelFormer + DGR obtaining the best RMSE of 2.041. These results are particularly noteworthy given the challenging nature of KITTI dataset, known for its diverse and dynamic outdoor scenes. The consistent improvements across different base models when integrated with DGR highlight the module's adaptability and effectiveness in various contexts. Figure 8 demonstrates the visualization results on the KITTI dataset. The incorporation of the DGR module leads to more precise object boundaries and a clearer representation of the depth differences between objects at varying distances. This improvement allows for more accurate estimation of contours for distant objects like trees and people, as well as cars. In Figure~\ref{fig:KITTI}(a), a red box highlights two signposts next to a white car, with one being farther away than the other. Figure~\ref{fig:KITTI}(b) reveals that after adding the DGR module, the depth information of these two signposts is predicted more accurately.

\iffalse
\subsection{Quantitative Evaluation Results}
We quantitatively evaluate the contribution of the DGR module to the Transformer's monocular depth estimation task on NYU-Depth-V2 dataset. The results are shown in Tables 2 and 3. DORN and Yin et al. are CNN-based models, and their performance on all evaluation metrics is slightly inferior compared to Transformer-based models. After incorporating DGR, all models' performance improved, especially PixelFormer + DGR, which outperformed all other models. This further underscores the importance of depth gradients in depth estimation tasks and highlights the superiority of the DGR module.
\fi

\paragraph{Loss Function} We further evaluated the performance of the PixelFormer + DGR model trained with different loss functions, as shown in Table~\ref{tab3}. When trained solely with the $L_{\mathrm{MSE}}(P,Q)$ loss function, the model achieved commendable performance on Abs Rel, RMS, and Log10, with a $\delta_1$ accuracy of 0.931. This suggests that the mean squared error loss already provides a stable optimization target for the model, enabling accurate depth prediction in most cases. When trained solely with the $L_{\mathrm{OTDL}}$ loss function, the model's performance was similar to that trained only with the mean squared error loss, hinting at a potential complementary relationship between the two. When combining both loss functions, the model achieved the best performance on all evaluation metrics.

\begin{table}[]
	
	\centering
	\resizebox{1\columnwidth}{!}{%
			\begin{tabular}{cc|cccc}
			\hline
			\multicolumn{2}{c|}{Loss function} & \multicolumn{1}{c}{\multirow{2}{*}{Abs Rel$\downarrow$}} & \multicolumn{1}{c}{\multirow{2}{*}{RMS$\downarrow$}} & \multicolumn{1}{c}{\multirow{2}{*}{${\rm Log}_{10}\downarrow$}} & \multicolumn{1}{c}{\multirow{2}{*}{$\delta_1\uparrow$}} \\ \cline{1-2}
			\multicolumn{1}{c|}{$L_{\mathrm{MSE}}(P,Q)$} & \multicolumn{1}{c|}{$L_{\mathrm{OTDL}}$} & \multicolumn{1}{c}{} & \multicolumn{1}{c}{} & \multicolumn{1}{c}{} & \multicolumn{1}{c}{} \\ \hline
			\checkmark& & 0.088 & 0.320 & 0.038 & 0.931 \\
			& \checkmark & 0.088 & 0.319 & 0.037 & 0.934 \\
			\checkmark	& \checkmark & \textcolor{blue}{\textbf{0.086}} & \textcolor{blue}{\textbf{0.310}} & \textcolor{blue}{\textbf{0.036}} & \textcolor{blue}{\textbf{0.937}} \\ \hline
		\end{tabular}%
	}
 
	\caption{Performance of models trained with different loss functions on NYU-Depth-V2.}
 
 \label{tab3}
\end{table}

\section{Conclusion}

In this study, we delved into the application and challenges of the Transformer architecture in monocular depth estimation. Through visual comparisons between Transformer and CNN models, we observed the advantages of the Transformer in depth cue selection and also identified its potential issues, such as unnatural depth jumps in certain smooth areas. To enhance the performance of the Transformer in monocular depth estimation tasks, we introduced two novel approaches: the Depth Gradient Refinement (DGR) module and the Optimal Transport Depth Loss (OTDL). The DGR module leverages higher-order derivatives to capture intricate variations in depth maps, especially in regions where object edge intensities change rapidly. This not only enhances the continuity of the depth map but also heightens its sensitivity to object edges. On the other hand, OTDL offers a more refined loss function for the Transformer, emphasizing the preservation of depth distribution variance and continuity. When combined with the state-of-the-art Transformer-based models, our methods not only surpassed existing performance benchmarks but also provided deeper visual and conceptual insights into the working mechanism of the Transformer. This lays a solid foundation for future research and applications of Transformers in monocular depth estimation. 

{
    \small
    \bibliographystyle{ieeenat_fullname}
    \bibliography{main}
}


\end{document}